\documentclass[10pt,twocolumn,letterpaper]{article}

\usepackage{wacv}              % To produce the CAMERA-READY version
\usepackage[utf8]{inputenc} % allow utf-8 input
\usepackage[T1]{fontenc}    % use 8-bit T1 fonts
\usepackage{url}            % simple URL typesetting
\usepackage{booktabs}       % professional-quality tables
\usepackage{amsfonts}       % blackboard math symbols
\usepackage{nicefrac}       % compact symbols for 1/2, etc.
\usepackage{microtype}      % microtypography
\usepackage[dvipsnames]{xcolor}         % colors

\usepackage{graphicx}       % images
\usepackage[outdir=./]{epstopdf}   % Needed to run on arXiv
\usepackage[dvipsnames]{xcolor}  % Colors for the author's feedback
\usepackage{multirow}
\usepackage{siunitx}        % Allow formatting in the tables. Some values are italics.
\usepackage{float}          % Prevents that new tables in separate files create new pages
\usepackage{comment}        % Allows to comment blocks of text
\usepackage{amsmath}
\usepackage{colortbl}
\usepackage[flushleft]{threeparttable}
\usepackage{pifont}         % Contains the red cross character
\usepackage{tabularx}
\usepackage{cuted}

\newcommand{\greentick}{\textcolor{green}{\ding{51}}}
\newcommand{\redcross}{\textcolor{red}{\ding{55}}}

\newcommand{\Ngen}{$N_{Gen}$}
\newcommand{\Naqa}{$N_{AQA}$}
\newcommand{\Nmqa}{$N_{MQA}$}
\newcommand{\MQA}{\textrm{MQA}}
\newcommand{\ratio}{$y_{\textrm{AQA}}$}
\newcommand{\precision}{$P_{\textrm{AQA}}$}
\newcommand{\recall}{$R_{\textrm{AQA}}$}

\definecolor{wacvblue}{rgb}{0.21,0.49,0.74}
\usepackage[pagebackref,breaklinks,colorlinks,allcolors=wacvblue]{hyperref}

\def\wacvPaperID{675} % *** Enter the WACV Paper ID here
\def\confName{WACV}
\def\confYear{2026}

\title{Cost Savings from Automatic Quality Assessment of Generated Images}

\author{Xavier Giro-i-Nieto
\quad\quad
Nefeli Andreou
\quad\quad
Anqi Liang
\\
Manel Baradad
\quad\quad
Francesc Moreno-Noguer
\quad\quad
Aleix Martinez
\\ 
\emph{Amazon}
}

\begin{document}

\maketitle
%\vspace{1cm}
\begin{abstract}
Deep generative models have shown impressive progress in recent years, making it possible to produce high quality images with a simple text prompt or a reference image. However, state of the art technology does not yet meet the quality standards offered by traditional photographic methods. For this reason, production pipelines that use generated images often include a manual stage of image quality assessment (IQA). This process is slow and expensive, especially because of the low yield of automatically generated images that pass the quality bar. The IQA workload can be reduced by introducing an automatic pre-filtering stage, that will increase the overall quality of the images sent to review and, therefore, reduce the average cost required to obtain a high quality image. We present a formula that estimates the cost savings depending on the precision and pass yield of a generic IQA engine. This formula is applied in a use case of background inpainting, showcasing a significant cost saving of 51.61\% obtained with a simple AutoML solution. 
\end{abstract}

\section{Motivation}

% The benefits of automatic image generation
The automatic production of images has experienced a rapid progress thanks to the broad adoption of deep neural networks in generative models learning. The current solutions based on diffusion~\cite{peebles2023scalable,chen2024pixartalpha} and auto-regressive~\cite{yuscaling,tian2025visual} models can produce high quality images, sometimes indistinguishable to the naked eye from their photographic counterparts. 
This represents an opportunity for industry because automatic image generation is orders of magnitude less expensive than a manual capture, which typically requires a photography studio and trained professionals.
Moreover, the time necessary to generate an image is virtually zero compared to a manual production, and the scale is basically limited by the available computational resources.
These advantages make generative deep learning an attractive technology for reducing costs and increasing the scale. %ing the scale of their production.

\begin{figure}[htb]
    \centering
    \includegraphics[width=0.45\textwidth]{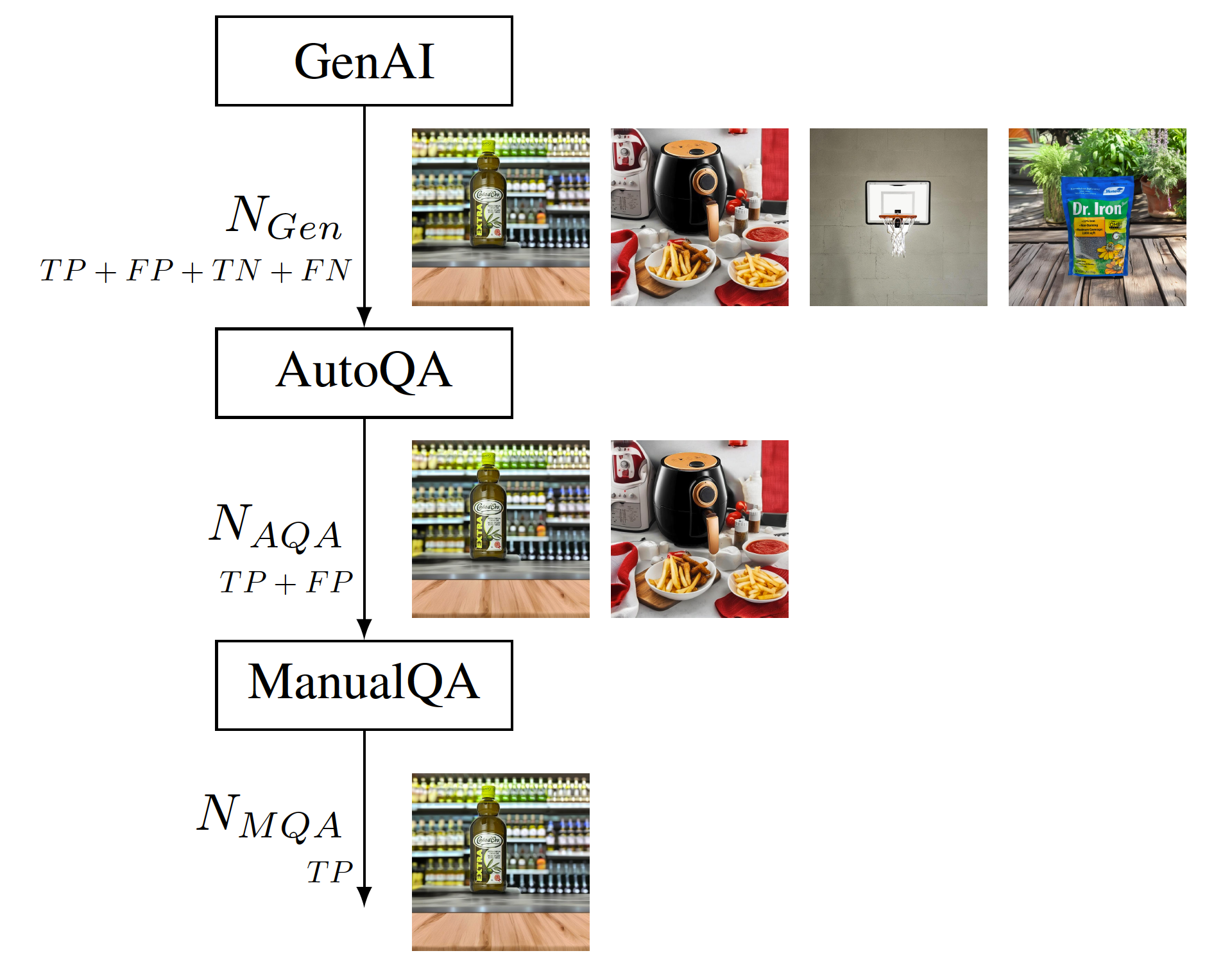}
    \caption{Vertical pipeline for image generation and quality assessment with sample images. $N_{Gen}$, $N_{AQA}$ and $N_{MQA}$ represent the number of images output after each block. TP/FP/TN/FN refer to true/false positive/negatives.}
    \label{fig:pipeline_vertical}
\end{figure}

\begin{figure*}[t]
    \centering
    \begin{tabular}{m{0.14\textwidth}m{0.14\textwidth}m{0.14\textwidth}m{0.14\textwidth}m{0.14\textwidth}m{0.14\textwidth}}
        \includegraphics[width=0.14\textwidth]{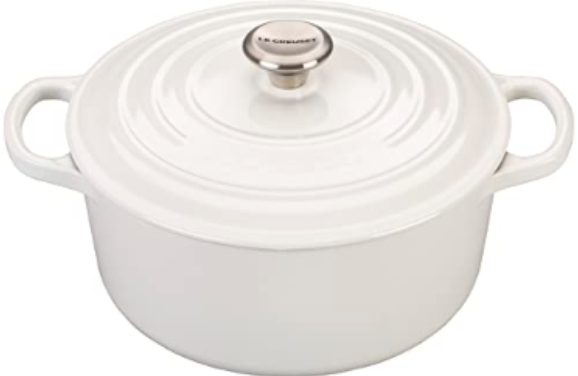}
        &
        \includegraphics[width=0.14\textwidth]{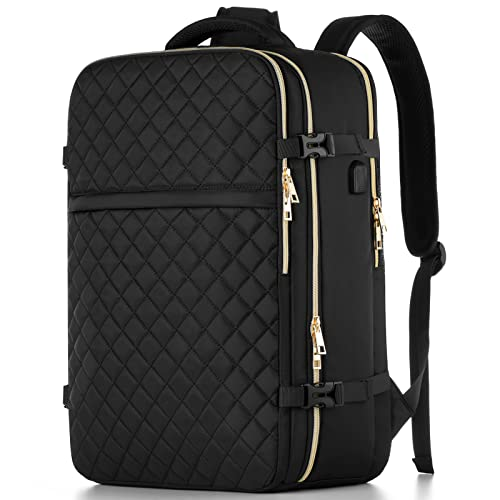}
        &
        \includegraphics[width=0.14\textwidth]{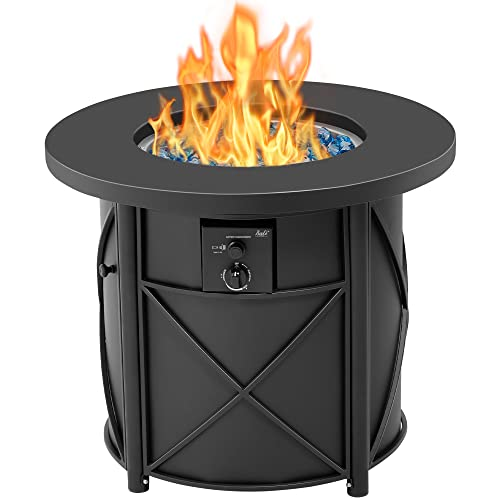}
        &
        \includegraphics[width=0.14\textwidth]{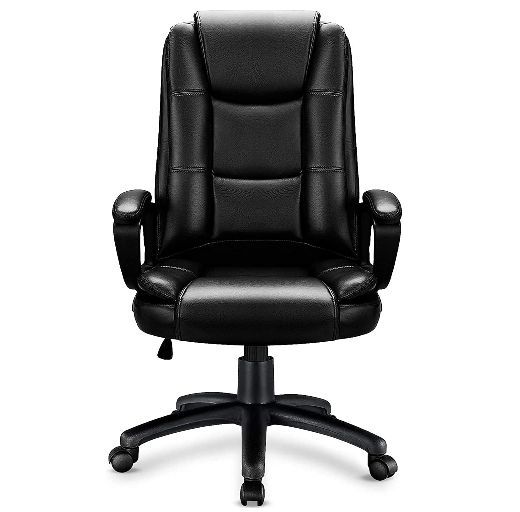}
        &
        \includegraphics[width=0.14\textwidth]{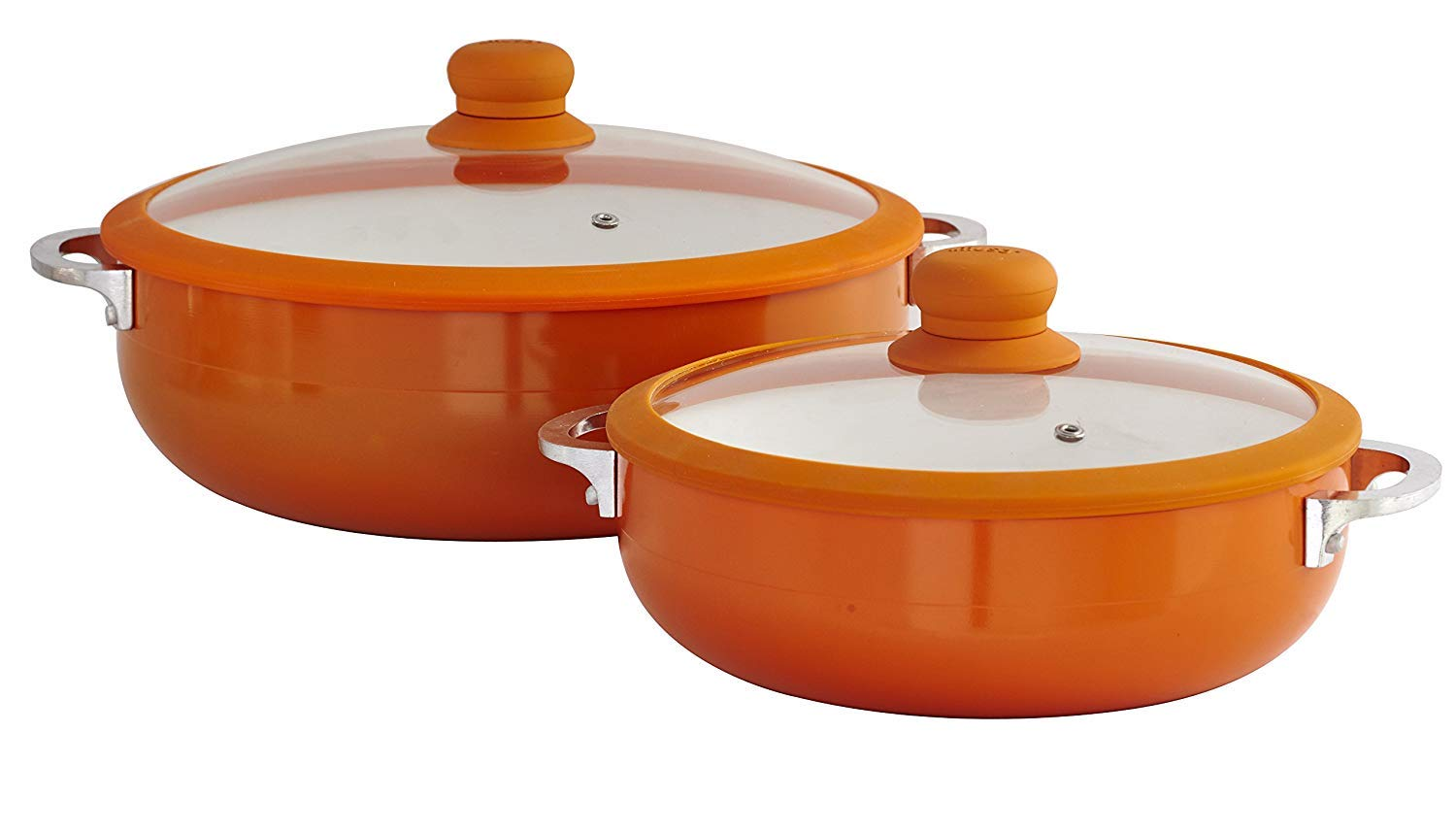}
        &        
        \includegraphics[width=0.14\textwidth]{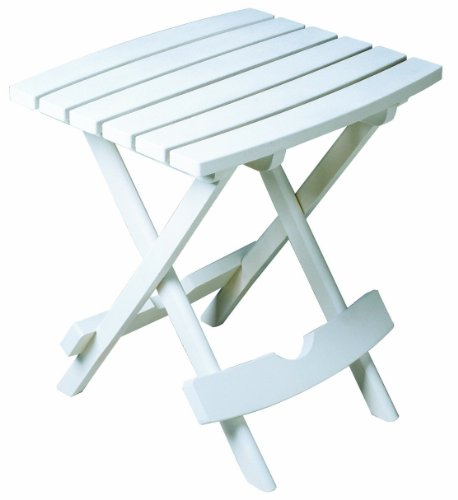}
        \\    
        \includegraphics[width=0.14\textwidth]{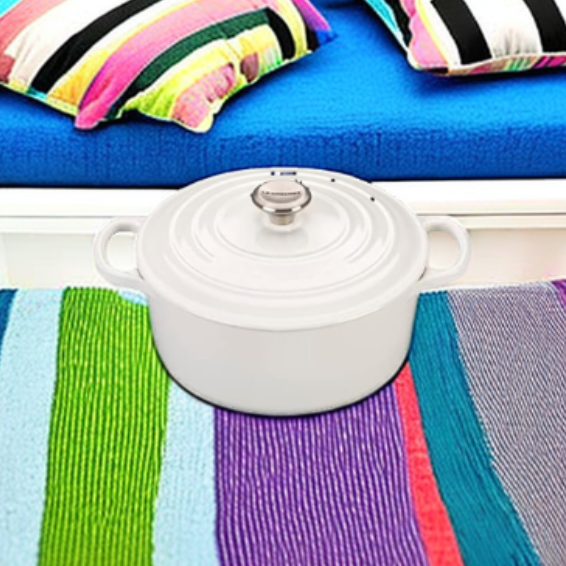}
        &
        \includegraphics[width=0.14\textwidth]{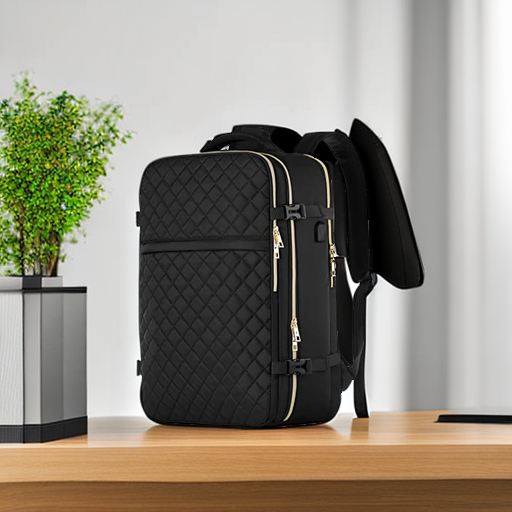}
        &
        \includegraphics[width=0.14\textwidth]{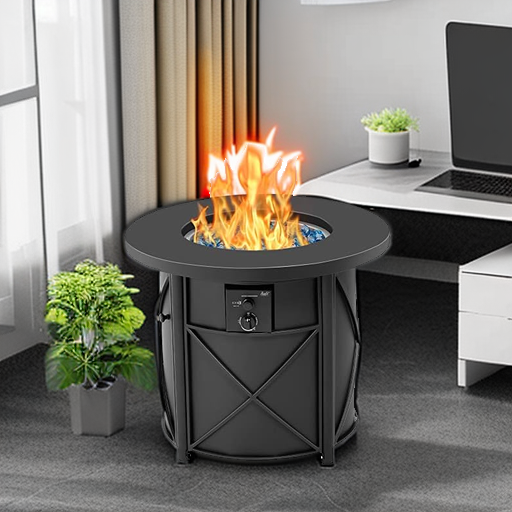}
        &
        \includegraphics[width=0.14\textwidth]{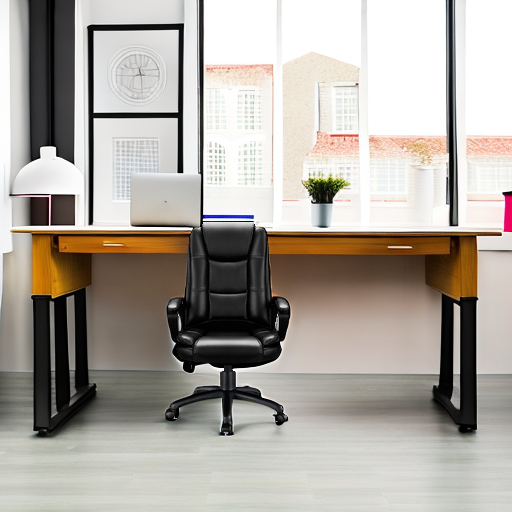}
        &
        \includegraphics[width=0.14\textwidth]{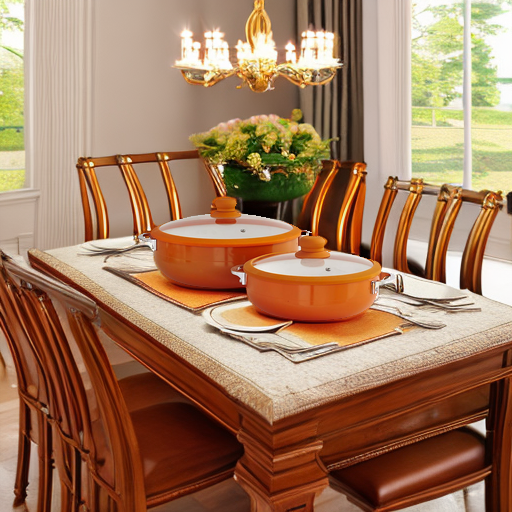}
        &        
        \includegraphics[width=0.14\textwidth]{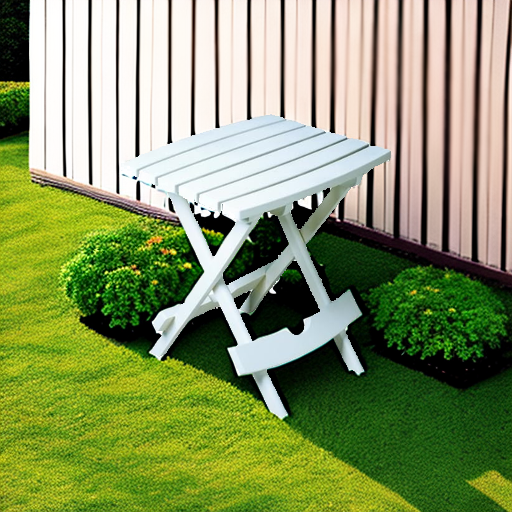}
        \\
        \centering\small\parbox{0.14\textwidth}{\centering Main Object Distortion}
        &
        \centering\small\parbox{0.14\textwidth}{\centering Main Object Extension}
        &
        \centering\small\parbox{0.14\textwidth}{\centering Misplaced Object}
        &
        \centering\small\parbox{0.14\textwidth}{\centering Scale Mismatch}
        &
        \centering\small\parbox{0.14\textwidth}{\centering Background \\Objects Distortion}
        &        
        \centering\small\parbox{0.14\textwidth}{\centering Background Structural Distort.}
    \end{tabular}
    \caption{Types of visual defects in background inpainting. Top: Reference object. Bottom: Generated image.}
    \label{fig:defects}
\end{figure*}

% The limitations of automatic image generation
However, current technology is not perfect, and may introduce noticeable artifacts in a significant portion of the generated images.
The nature of these defects may be diverse, whether directly visible on the generated images, or failing to follow the provided guidelines, which typically take the form of a textual prompt or a visual reference.
For example, in the \textit{inpainting} study case that we explore in this work, the goal is generating a realistic visual context for an object depicted over a white background.
%the study case presented in our experiments focuses in outpainted images, that is, that generate a background for a given segmented object.
This is a specially challenging problem, because not only the background must be perceptually realistic, but it must also be semantically coherent with the provided reference object.
Figure~\ref{fig:defects} shows six different types of defects that are common in background inpainting.
The details of these defects are described in Table~\ref{tab:defects_examples}.
%, produced with different state of the art image generation models.
According to our studies with marketing professionals, existing technology offers low QA pass yield for this task: only 1.87\% with the model considered in this work, based on Stability AI SDXL~\cite{podell2023sdxl}. 
%Supplementary  , and around 16\% with a solution based on Flux~\cite{flux2024}, which is analyzed in the supplementary material.

% How to mitigate the limitations of automatci image generation with AutoQA
Commercial applications require high quality images, so an additional filtering is currently needed before delivering  automatically generated images. 
A first solution is relying only on human annotators trained to detect the most common defects, an approach we refer to as \textit{ManualQA}. %, for \textit{manual quality assessment}.
This approach is simple in terms of technology, because it only requires a user interface to collect the annotations. %, and will typically predict the images that pass the quality bar with high precision . 
However, ManualQA also presents challenges in terms of calibrating the annotators with the product requirements and, more importantly, scales very poorly in terms of time and costs.
Another approach is introducing a computer vision system capable of detecting the anomalies in the generated images, which we will refer as \textit{AutoQA}. % for "automatic quality assessment".
This is a scalable solution whose speed is only limited by the allocated computational resources. 
On the other hand, existing computer vision systems offer a lower precision than human annotators.

% \begin{table}[htbp]
%     \centering
%     \begin{tabular}{ll}
%         \toprule
%         \textbf{Type of defect} & \textbf{Description} \\
%         Main Object Distortion  & Black dots on the pan cover. \\
%         Bg. Objects Distortion  & Unrealistic chair backs. \\        
%         Objects Extensions  & Straps added to the backpack. \\        
%         Misplaced Object & The firepit is an outdoor object but is placed in an indoor environment. \\
%         Scale Mismatch  & The chair is much smaller than the table. \\        
%         Bg. Structure Distortion  & The wall becomes misaligned behind the table. \\                
%         \bottomrule
%     \end{tabular}
%     \caption{Defect categories and their corresponding evaluation prompts}
%     \label{tab:defects_examples}
% \end{table}

% \usepackage{tabularx} % Add this to your preamble

\begin{table}[htbp]
    \centering
    \begin{tabularx}{0.47\textwidth}{lX}
        \toprule
        \textbf{Type of defect} & \textbf{Description} \\
        \midrule
        Main Object Distortion  & Black dots on the pan cover. \\
        Main Object Extension  & Straps added to the backpack. \\        
        Misplaced Object & The firepit is an outdoor object but is placed in an indoor environment. \\
        Scale Mismatch  & Chair is much smaller than table. \\        
        Bg. Objects Distortion  & Unrealistic chair backs. \\                
        Bg. Structural Distortion  & Misaligned wall behind the table. \\                
        \bottomrule
    \end{tabularx}
    \caption{Detailed descriptions of the defects in Figure~\ref{fig:defects}.}
    \label{tab:defects_examples}
\end{table}

In this work, we consider the hybrid pipeline depicted in Figure~\ref{fig:pipeline_vertical}, where ManualQA will only process those images that pass AutoQA.
This scheme aims at increasing the pass yield of ManualQA by pre-filtering the generated images with AutoQA.
As a result, the final cost of generating a given amount of images is lower thanks to the reduced amount of reviews needed.

Our contributions are three-fold: 
(a) a novel IQA task for images generated by background inpainting, focused on the popular application of placing objects in generated contexts
(b) a closed formula that determines the cost savings of introducing an AutoQA block in an image generation system, and 
a (c) a case study based on a zero-shot VLM and AutoML which, in their best set up, obtain financial cost savings of 51.61\%.
\section{Related work}

Image evaluation methodologies in the literature can be broadly categorized by 
(1) the rating of individual images~\cite{ssimpsnr} or comparative ranking across image sets~\cite{Shi:2020} (2) by the availability of reference images that are used for comparison (full-reference methods~\cite{fsim}, reduced-reference methods~\cite{Rehman:2012}, or no-reference methods~\cite{Mittal:2012}, or (3) by the main features used including traditional image features~\cite{ssimpsnr,Chandler:2007} or deep features (such as representations extracted from pre-trained models)~\cite{xu2023imagereward,hessel2021clipscore,zhang2018perceptual,hu2023tifa}. Our work concentrates on individual image quality assessment in a reference-free manner. Image Quality Evaluation can span multiple dimensions, including image realism~\cite{fid,inceptionscore,kid,zhang2018perceptual}, text-image alignment~\cite{xu2023imagereward,yarom2023you,hessel2021clipscore}, image aesthetics~\cite{yi2023towards,Yang_2022_CVPR,sheng2023aesclip} and human preferences~\cite{wu2023humana,wu2023humanb,zhang2024learning,kirstain2023pick}. Our work considers the evaluation of realism in generated imagery.

\noindent \textbf{Automated Image Quality Assessment (IQA)}. 
Automatic Image Quality Assessment has become a critical block when generating images, and multiple benchmarks and metrics have been proposed~\cite{ssimpsnr,hartwig2025surveyqualitymetricstexttoimage}.
When comparing an distorted image with its reference, PSNR and MSE perform pixel-level comparisons, while SSIM~\cite{ssimpsnr} captures perceptual changes through structural information by comparing luminance, contrast, and structure. 
~\cite{cheon2021perceptual} proposed an Image Quality Transformer (IQT) to measure the perceptual quality between an input image pair of distorted and reference images. Given the fact that reference images are not always available,  
\cite{you2021transformer, conde2022conformer} explore the performance of transformer-based no-Reference image quality assessment (NR-IQA). Along the NR-IQA research line, ~\cite{Wang_2022_CVPR} proposed an algorithm based on the Swin Transformer ~\cite{liu2021swin} with fused features from multiple stages to better predict image quality. 

%\textcolor{purple}{
\noindent \textbf{Image Quality Assessment with VLMs}
The surprising understanding capabilities of large 
 Vision-Language Models (VLMs) have been explored for image quality assessment. CLIPScore~\cite{hessel2021clipscore} uses CLIP embeddings for evaluating text-image alignment. Building upon this foundation, ImageReward~\cite{xu2023imagereward} introduced a reward model fine-tuned on human preferences to better capture image quality aspects. PickScore and HPS~\cite{kirstain2023pick,wu2023humana} align better with subjective human judgement in the evaluation.
Similar to GenomeBench~\cite{Corneanu_2024_WACV} which proposes a benchmark to evaluate the quality and text alignment of generated images through a series of questions and human annotations, recent works have leveraged large Vision-Language Models (VLMs) for nuanced assessment of image realism through visual question answering (VQA)~\cite{antol2015vqa}. Foundational models can engage in detailed dialogues about image artifacts and realism. InstructBLIP~\cite{dai2023instructblip} demonstrates how instruction-tuning of VLMs can enable more targeted assessment of specific quality aspects. TIFA~\cite{hu2023tifa} leverages VQA to provide interpretable assessment of text-image alignment. Cho~\cite{cho2024davidsonian} built the set of questions using Davidsonian graphs. Gecko~\cite{wiles2024gecko} proposed a fine-grained classification of defects in prompt alignment, named \textit{skills}. 
VQAScore~\cite{lin2024evaluating} showed better performance of a VLM than the CLIPScore for IQA.

% Make the colors in this plot color-blind friendly
% Rearrange the columns so that the order is internal\distortion, object\extension,  misplaced\object, scale\mismatch, bg. objects\distortion, bg. structural\distortion

\begin{figure*}[htb]
    \centering
    \includegraphics[width=0.7\textwidth]{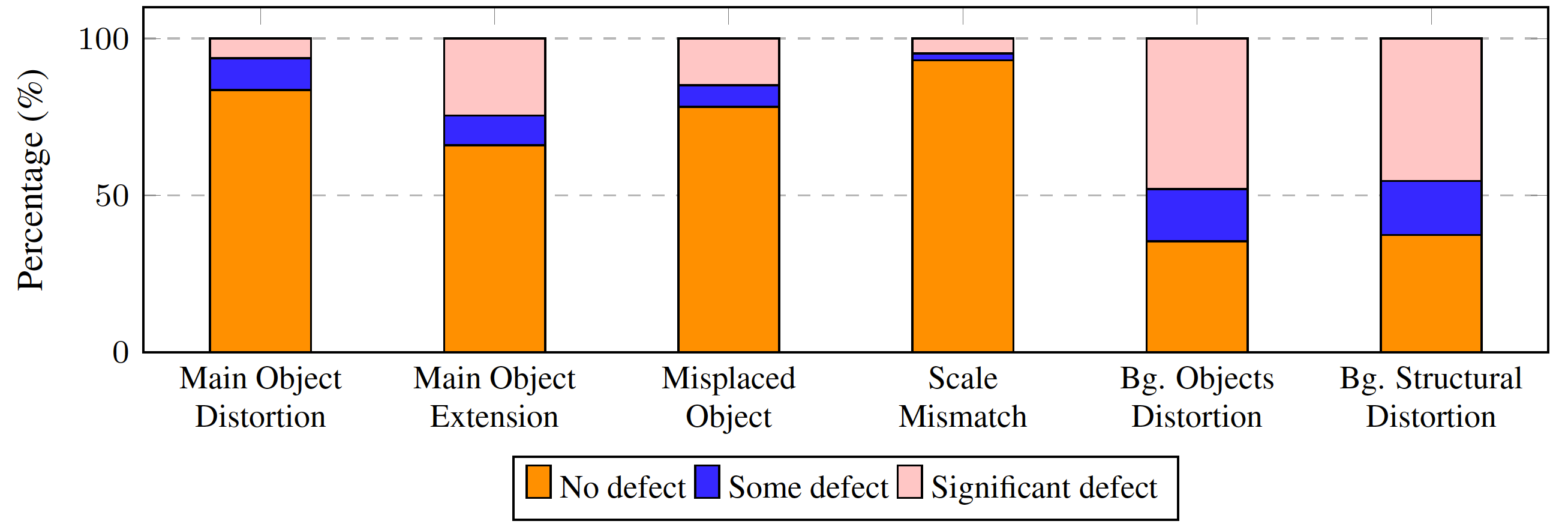}
    \caption{Distribution of defect severity across the six types of coarse defects. Each of the 8,304 generated images was manually labeled by a human annotator according to a scale of three levels: \textit{No defect}, \textit{Some defect}, or \textit{Significant defect}. This figure does not consider agreement among annotators}
    \label{fig:defect_distribution}
\end{figure*}

\section{Cost Estimation}
\label{sec:cost_estimation}

We present a formula that estimates the cost savings from introducing AutoQA for automatic image generation, following the pipeline in Figure~\ref{fig:pipeline_vertical}. The QA classification systems operate with binary classes: \textit{Clean} (denoted with \greentick) and \textit{Defect} (denoted with \redcross). An image generation engine (GenAI) produces \Ngen images that are first evaluated by the AutoQA block.
Only the \Naqa samples that pass AutoQA are reviewed in ManualQA.
The result are \Nmqa high quality images that passed both QA controls.

We relate the volume of images after each block (\Ngen, \Naqa, \Nmqa) through the two metrics that characterize the blocks in Figure~\ref{fig:pipeline_vertical}: the \textit{yield} of images that pass AutoQA (\ratio) and ManualQA ($y_{MQA}$), and the \textit{precision} of the image generator ($P_{Gen}$) and AutoQA engine (\precision).
The two metrics are interpretable and facilitate the estimation of financial costs in industrial scenarios.

\subsection{Volume of Images}
\label{sec:volume}

As introduced in Figure~\ref{fig:pipeline_vertical}, we develop a formulation that considers \textit{Clean} (\text{\greentick}) as the positive class of our task, and uses the standard notation of TP/FP/TN/FP for referring to true/false positives/negatives.
Based on these definitions, we can relate \Nmqa with \Naqa as

\begin{equation}
\scalebox{0.8}{$
\begin{split}
N_{MQA} &= TP = \frac{TP+FP}{TP+FP} \cdot \frac{TP}{TP} \cdot TP = \\
&= (TP+FP) \cdot  \frac{TP}{TP+FP} \cdot \frac{TP}{TP} = \\
&= N_{AQA} \cdot P_{AQA}(\text{\greentick}) = N_{AQA} \cdot y_{MQA}, 
\end{split}
$}
\label{eq:N_ManualQA_AutoQA}
\end{equation}

where $P_{AQA}(\text{\greentick})$ is the precision of the AutoQA module for the \textit{Clean} class.
Notice that $P_{AQA}(\text{\greentick})$ can also be interpreted as the ManualQA yield, $y_{MQA}$, because we consider ManualQA as the source of true predictions.

Similarly, we can relate \Nmqa~with \Ngen~as

\begin{equation}
\scalebox{0.8}{$
\begin{split}
N_{MQA} &= TP = \frac{TP+FP+TN+FN}{TP+FP+TN+FN} \cdot \frac{TP+FP}{TP+FP}  \cdot TP = \\
&= (TP+FP+TN+FN) \frac{TP+FP}{TP+FP+TN+FN} \frac{TP}{TP+FP}  = \\
&=  N_{Gen} \cdot y_{AQA} \cdot P_{AQA}(\text{\greentick}) = N_{Gen} \cdot y,
\end{split}
$}
\label{eq:N_ManualQA_GenAI}
\end{equation}

where \ratio is the AutoQA yield.
We also define the overall yield $y = y_{AQA} \cdot P_{AQA}(\text{\greentick}) = \frac{TP}{TP+FP+TN+FN}$ to simplify the formulation in future equations.

\subsection{Total Cost}

The total cost of a pipeline with AutoQA can be derived by weighting the unitary cost ($c_i$) of processing the images in each of the three stages $i \in \lbrace Gen, AQA, MQA \rbrace$, and summing their results as

\begin{equation}
\scalebox{0.8}{$
\begin{split}
C_{Total} &= N_{Gen} \cdot c_{Gen} + N_{Gen} \cdot c_{AQA} + N_{AQA} \cdot c_{MQA} = \\
&= \frac{N_{MQA}}{y} c_{Gen} + \frac{N_{MQA}}{y } c_{AQA} + \frac{N_{MQA}}{P_{AQA}(\text{\greentick})} c_{MQA} = \\
&= N_{MQA} \cdot \left( \frac{c_{Gen}}{y} + \frac{ c_{AQA}}{y} + \frac{c_{MQA}}{P_{AQA}(\text{\greentick})}  \right).
\end{split}
$}
\label{eq:autoqa_cost}
\end{equation}

\subsection{Cost Savings}

The cost savings are estimated by comparing $C_{TOTAL}$ with the costs of obtaining the same amount of images without AutoQA, which is

\begin{equation}
\scalebox{0.8}{$
\begin{split}
C_{Baseline} &= N_{Gen} \cdot c_{Gen} + N_{Gen} \cdot c_{MQA} =\\ &=N_{Gen} \left( c_{Gen} + c_{MQA} \right) =\\
&=\frac{N_{MQA}}{P_{Gen}(\text{\greentick})} \cdot \left( c_{Gen} + c_{MQA} \right) = \\
&= N_{MQA} \cdot \left( \frac{c_{Gen}}{P_{Gen}(\text{\greentick})} + \frac{c_{MQA}}{{P_{Gen}(\text{\greentick})}} \right),
\\
\end{split}
$}
\label{eq:manualqa_cost}
\end{equation}
where $P_{Gen}(\text{\greentick})$ is the precision of the GenAI technology that produces the images, that is, the ManualQA yield.

The final cost savings $\Delta C$ can be calculated by substracting the total costs defined in Equation~\ref{eq:autoqa_cost} from the baseline costs in Equation~\ref{eq:manualqa_cost}, as 

\begin{equation}
\scalebox{0.8}{$
\begin{split}
\Delta C &= C_{Baseline} - C_{Total} = \\
&= N_{MQA} \cdot \left( \frac{c_{Gen}}{P_{Gen}(\text{\greentick})} + \frac{c_{MQA}}{P_{Gen}(\text{\greentick})} - \right. \\
&\left. - \frac{c_{Gen}}{rP_{AQA}(\text{\greentick})} - \frac{c_{AQA}}{y} - \frac{c_{MQA}}{P_{AQA}(\text{\greentick})} \right) = \\
&= N_{MQA} \left( \frac{rP_{AQA}(\text{\greentick}) - P_{Gen}(\text{\greentick})}{y \cdot P_{Gen}(\text{\greentick})}c_{Gen} - \right. \\
&\left. - \frac{1}{y}c_{AQA}  + \frac{P_{AQA}(\text{\greentick}) - P_{Gen}(\text{\greentick})}{P_{AQA}(\text{\greentick}) \cdot P_{Gen}(\text{\greentick})}c_{MQA} \right)
\end{split}
$}
\label{eq:cost_savings}
\end{equation}

\section{Experiments}

We apply the cost savings formula from Equation~\ref{eq:cost_savings} in a realistic case for e-commerce: the generation of a virtual context from images that depict products over a white background.
%e-commerce imaginery of products over a white background are modified to show the object in context. 
%generated images add a background to a given reference object on a white background.
This problem is technically referred as background \textit{inpainting}, and the resulting imagery \textit{lifestyle} images. %, and requires a binary mask that segments the object, which was also given
%This is a relevant application in e-commerce, where images of the products are often available over a white background, but lifestyle images that show the product in a context are missing.
We adopt state of the art technical solutions for both generating and evaluating the quality of the generated images. %, so that cost savings are realistic.
% Prompt for Claude
% Create a Latex table whose cells contain the values of columns "precision_defect" and "ratio_defect". The table must focus only on the rows where "prompt version" equals 1. Each row will contain the "defect_id", and the columns will correspond to "# samples", "autogluon", "novapro", "oracle" and "random". Under each model, we should have two columns: one for "ratio_defect" and another one for "precision_defect". Do not include any vertical line. The latex label for the table must be "tab:autogluon_vs_novapro". Show the headers in bold. For the Oracle column, only show "ratio",  and discard "prec". Move this column between "# samples" and "Autogluon". Still keep the "ratio" label under the "Oracle" header. Multiply the values under columns "%" per 100. 

\begin{table*}[htb]
\centering
\begin{tabular}{l c c cc cc cc}
\hline
\multirow{2}{*}{\textbf{Defect type}} & \multirow{2}{*}{\textbf{\# img}} & \multicolumn{1}{c}{\textbf{Oracle}} & \multicolumn{2}{c}{\textbf{Autogluon 1.2}} & \multicolumn{2}{c}{\textbf{Nova Pro 1.0}} & \multicolumn{2}{c}{\textbf{Random 0.5}} \\
\cmidrule(lr){3-3} \cmidrule(lr){4-5} \cmidrule(lr){6-7} \cmidrule(lr){8-9}
 & & \ratio & \ratio & \precision(\redcross)$\uparrow$ & \ratio & \precision(\redcross)$\uparrow$ & \ratio & \precision(\redcross)  $\uparrow$ \\
\hline
Main Object Distortion & 199 & 0.96 & \textcolor{red}{1.00} & \textcolor{red}{0.000} & 0.98 & \textcolor{red}{0.000} & 0.46 & 0.037 \\
Main Object extension & 231 & 0.73 & \textbf{0.82} & \textbf{0.561} & \textcolor{red}{0.97} & 0.500 & 0.52 & 0.312 \\
Product Placement & 212 & 0.83 & \textcolor{red}{0.97} & \textbf{0.571} & \textbf{0.87} & 0.382 & 0.44 & 0.144 \\
Scale Mismatch & 199 & 0.97 & \textbf{0.98} & 0.667 & \textcolor{red}{0.80} & \textcolor{red}{0.022} & 0.52 & 0.021 \\
Bg. Objects Distortion & 254 & 0.68 & 0.77 & \textbf{0.508} & \textcolor{red}{0.82} & \textcolor{red}{0.370} & 0.56 & 0.312 \\
Bg. Structural distortion & 225 & 0.74 & \textbf{0.80} & \textbf{0.444} & \textcolor{red}{0.98} & \textcolor{red}{0.105} & 0.48 & 0.284 \\
\hline
\end{tabular}
\caption{Comparison of yield, \ratio, and precision for the \textit{Defect} class, \precision(\redcross), for the two considered AutoQA technologies. 
Column \textit{\# img} indicates the amount of test images considered when assessing each defect.
Column \textit{Oracle} represents the defect rate of images that a perfect AutoQA system would detect.
%, and serves as a reference to assess AutoGluon and Nova Pro, who should produce a similar rate with a \precision(\redcross) near $0.5$.
Values in \textbf{bold} highlight which of the two technologies perform better to detect each defect type, by considering: (a) a \ratio~ similar to the Oracle's, and (b) a \precision $\approx 0.5$.
Values in \textcolor{red}{red} indicate failure cases, due to \precision~ similar to the \textit{Random} case, and/or \ratio~ very different from the Oracle's.
}

% the provides the ratio of defects over the total amount of test images, according to the expert annotations.  
% Values in \textcolor{red}{red} highlight the cases where AutoQA fails, which corresponds to very different \ratio(\redcross) from the oracle's, and/or \precision(\redcross)$ \approx 0$.}
\label{tab:autogluon_vs_novapro}
\end{table*}

\subsection{Dataset}

% Description of the images
Our experiments were conducted with a dataset of 8,304 images of 195 different product types. %product types.
The images were generated with a latent diffusion-based model open-sourced by Stability AI (SDXL)~\cite{podell2023sdxl}, which was adapted with ControlNet~\cite{Zhang_2023_ICCV} to solve the inpainting task. % with the DeepCatalog dataset.
We worked with a team of photography experts to define six categories of the possible defects, depicted in Figure \ref{fig:defects}. 

% Description of the labels
Professional annotators from an external vendor were trained to provide a numerical rating on a scale from 1 (\textit{No defect}) to 3 (\textit{Significant defect}).
Each image was labeled by 3 different annotators from a pool of 21.
The average amount of annotations per worker was 1256.38, with a maximum of 1782 and a minimum of 815.
The label distribution of the collected labels is presented in Figure~\ref{fig:defect_distribution}. The plots show how the low-level distortions on the background are the most common defects (\textit{Bg. Objects Distortion} and \textit{Bg. Structural Distortion}), while the mismatch of scale between the provided reference object and the rest of the scene is the most rare defect (\textit{Scale Mismatch}). 

\begin{table}[htbp]
\centering
\begin{tabular}{lc}
\hline
\textbf{Defect Type} & \textbf{Agreement Rate $\uparrow$} \\
\hline
Main Object Distortion & 0.6525 \\
Main Object Extension & 0.5425 \\
Product Placement & 0.6653 \\
Scale Mismatch & 0.8582 \\
Bg. Objects Distortion & 0.3868 \\
Bg. Structural Distortion & 0.2982 \\
\hline
\textbf{Any Defect} & \textbf{0.4579} \\
\hline
\end{tabular}
\caption{Annotator Agreement Rates by type of defect, and also in the \textit{Any Defect} case. This later case reflects the final IQA task, where a binary decision must be made between \textit{Clean}(\text{\greentick}) or \textit{Defect}(\redcross).}
\label{tab:complete_agreement_rates}
\end{table}

%The experiments reported in the following sections do not consider all images, only those where all annotators agreed.
The following sections report experiments based on different subsets of these annotated data.
In each experiment, we only used samples where all annotators agreed in the labels under study.
This way, metrics are more reliable and avoid the problem of low inter-annotator agreement, which is depicted in Table~\ref{tab:complete_agreement_rates}.
Notice how obtaining alignment among humans for IQA task is very challenging, despite providing detailed instructions to annotators.
In addition, the very few images where all annotators agreed on a rating of 2 (some issue), were also discarded.
\subsection{AutoQA}

Two technologies were considered to serve as AutoQA: a classic AutoML solution, and an off-the-shelf large visual language model (VLM). 

\noindent \textbf{AutoML} solutions consider different machine learning techniques to solve a task defined by an annotated dataset.
AutoML allows a quick development and provides solid baselines.
In our work, we adopt AutoGluon~\cite{tang2024autogluon}, which leverages the \textit{timm} model zoo of deep learning architectures for image analysis~\cite{rw2019timm}.
Under the hood, AutoGluon trains a variety of different models, uses bagging (bootstrap aggregation) to train them, and considers different architecture through stack-ensembling these models.
In our use case, AutoGluon automatically opted for a solution based on an ensemble of fine-tuned vision transformers.

\noindent \textbf{Large Visual Language Models (VLMs)}
are generative models pretrained with a very large dataset, typically, of Internet scale. 
They have shown promising results in a diversity of multimodal tasks, among them, visual-question answering (VQA).
We initially considered two VLMs for image quality assessment formulated as VQA, Amazon Lite and Pro 1.0~\cite{nova2024amazon}.
After an initial comparison between them, we opted for Nova Pro only, given the poorer performance of Amazon Lite for IQA.
The VLMs were conditioned with a prompt specific for each defect type, requesting an answer in the same scale as the human annotators: 1 for \textit{No defect}, 2 for \textit{Some defect}, and 3 for \textit{Significant defect}.

The VLMs were queried with the question prompts described in Table~\ref{tab:defects_prompts}.
These prompts were manually designed to detect each type of defect, and some of the defects were further divided in more fine-grained categories. % to obtain more precise descriptions.
The VLM was modulated with the role prompt presented in Table~\ref{tab:system_info}.

\begin{table*}[htbp]
    \centering
    \begin{tabular}{>{\raggedright\arraybackslash}p{3cm}>{\raggedright\arraybackslash}p{3.5cm}>{\raggedright\arraybackslash}p{7cm}}
        \toprule
        \textbf{Coarse defect} & \textbf{Detailed defect} & \textbf{Prompt} \\
        \midrule
        \multirow{3}{=}{\raggedright Main Object Distortion} 
            & Surface texture & Focus on the surface of the \{object\_class\}. Is there any distortion on its texture? \\
            & Color blending & Can you see weird color blending at its contours? \\
            & Structural distortion & Is there any structural distortion in the \{object\_class\}? \\
        \midrule
        \multirow{2}{=}{\raggedright Main Object Extension}
            & Product extension & Does the \{object\_class\} present a realistic shape? Compare the shape of the \{object\_class\} in the first generated image to the reference image and its segmentation mask. Make sure that the \{object\_class\} did not grow in extension when the background was generated \\
            & Product attached & Is there any other object attached to the \{object\_class\}? If so, is this attachment common and natural? \\
        \midrule
        \multirow{5}{=}{\raggedright Misplaced Object}
            & Objects layout & What objects appear in the scene? Are their relative positions natural? \\
            & Floating objects & Look at the \{object\_class\}. It must be standing on a surface. Otherwise, consider that it is floating, which is a severe issue. \\
            & Matching location & In which locations is the normally found? Does the context in the image represent one of these probable locations? \\
            & Functional location & Where is the \{object\_class\} located? Does it appear in a proper functional location? \\
            & Rich background & How is the background around the \{object\_class\}? The background must contain rich semantic and be aesthetically appealing. A solid or uniform background is not acceptable. \\
        \midrule
        Scale Mismatch
            & Scale mismatch & There is an anomaly in the size of the \{object\_class\} compared to the rest of objects in the scene. True or false? \\
        \midrule
        Background Objects Distortion 
            & Objects distortion & What objects appear in the image? Is there any distortion in any of them? \\
        \midrule        
        \multirow{2}{=}{\raggedright Background Structural Distortion}
            & General & Is there any structural distortion in the scene? \\
            & Because occlusion & Is the background behind the \{object\_class\} realistic? Make sure that there are no discontinuities in the generated background because of the occlusion of the \{product\_type\} \\
        \bottomrule
    \end{tabular}
    \caption{Question prompts organised in a hierarchy of coarse and detailed defects.}
    \label{tab:defects_prompts}
\end{table*}

\begin{table*}[htbp]
    \centering
    \begin{tabular}{p{2cm}p{12cm}}
        \textbf{Knowledge} & You are a vision-language assistant responsible for assessing the quality of synthetically generated images. 
        You have expertise in professional photography for e-commerce and design.
        You will receive a question and your task is to answer with the most appropriate score. \\
        \hline
        \textbf{Objective} & You are assessing the quality of a synthetically generated image depicting a \{product\_type\}. 
        This image is generated by adding a background to an image of a \{product\_type\}.
        The main \{product\_type\} is the primary object of the image.        
        The background is generated by a text-to-image model. \\
    \end{tabular}
    \caption{Text prompts for System knowledge and objective.}
    \label{tab:system_info}
\end{table*}

\subsubsection{Independent defects}
\label{ssec:independent}

Our first experiment explores which machine learning models can detect the considered defect types, so we focus our metrics on the \textit{Defect}(\redcross) class.
For this experiment, we built a dataset of images that present none or a single a single defect only, that is, we discarded images with multiple types of defects.

We compared AutoGluon 1.2 (AG) only with Amazon Nova Pro 1.0 (NP). 
Similarly how Nova Pro used a specific prompt for each type of defect, an independent AutoGluon binary classifier was trained for each defect type.
A different data subset was created for each defect, providing 90\% of the available images to AutoGluon for finetuning, and keeping the remaining 10\% for our tests with AutoGluon and Nova Pro.
By default, AutoGluon uses the a partition 90\% of the provided data for training, and the remaining 10\% for internal validation.
The AutoGluon models were left to train as much time as needed, which was in the order of minutes for each binary classifier.
Models were trained with a p3.2xlarge cloud desktop in Amazon Web Services (AWS) equipped with a single Tesla V100-SXM2 GPU with 16 GB of memory.

Table~\ref{tab:autogluon_vs_novapro} shows that, in four out of the six considered defect types, AutoGluon performs better than Nova Pro.
Only for the \textit{product placement} defect Nova Pro competes with Autogluon with a better \ratio(\redcross).
The \textit{main object distortion} defects are not captured by any of the two configurations.

% Create a Latex table from the given XLSX file. Label the table as "tab:cascade". 

\begin{table*}[htb]
\centering
\begin{tabular}{l ccc ccc}
\hline
& \multicolumn{3}{c}{\textbf{Defect(\redcross)}} & \multicolumn{3}{c}{\textbf{Clean(\greentick)}} \\
\cmidrule(lr){2-4} \cmidrule(lr){5-7}
\textbf{Configuration} & \ratio  & \precision $\uparrow$ & \recall $\uparrow$ & \ratio & \precision $\uparrow$ & \recall $\uparrow$ \\
\hline
Random (0.5) & 0.521 & 0.809 & 0.521 & 0.500 & 0.184 & 0.465 \\
%No AutoQA & 0.00 & 0.000 & 0.000 & 1.00 & \textcolor{blue}{0.187} & 1.000 \\
\hline
Cascade (AG) & 0.761 & 0.848 & 0.793 & 0.239 & 0.297 & 0.380 \\
Cascade (AG NP) & 0.847 & 0.823 & 0.858 & 0.153 & 0.241 & 0.197 \\
Single (AG)  & 0.882 & 0.835 & 0.916 & \textbf{0.118} & \textbf{0.400} & 0.211 \\
\hline
Oracle & 0.813 & 1.000 & 1.000 & 0.187 & 1.000 & 1.000 \\
\hline
\end{tabular}
\caption{Performance metrics on 380 test images for three AutoQA configurations.}
\label{tab:autoqa_sdxl}
\end{table*}

\subsubsection{All defects}
\label{ssec:all_defects}

Based on the results of detecting defects independently, we explore the performance of a full AutoQA system that would detect all defects but \textit{main object distortion}.

\begin{figure*}[!htb]
    \centering
    \includegraphics[width=\textwidth]{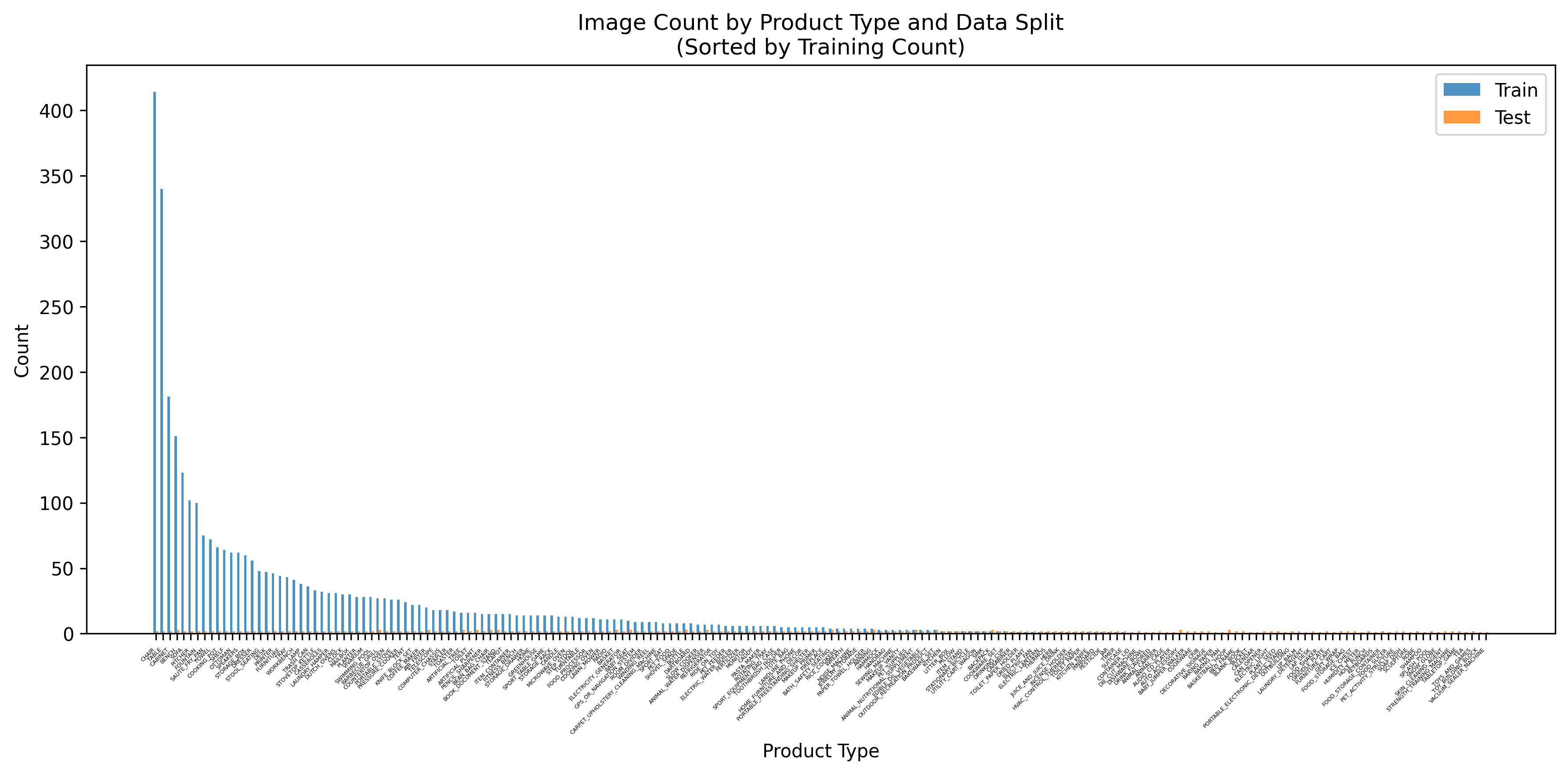}
    \caption{Histogram by object category of the train and test partition used to in the \textit{All Defects} experiment. }
    \label{fig:product_type_histogram_by_train}
\end{figure*}

% Dataset
The dataset in this section no longer satisfies the restriction of presenting a single defect.
In this case, the full dataset contains 3,802 images, and its test partition 380 samples with unbalanced binary labels. 
Only 18.7 \% of the test images are \textit{Clean}, and the rest contain one or more defects.
The test partition covers 192 different object categories, so most categories only have two samples.
On the other hand, the training partition used for AutoGluon only covers 140 of these 192 categories, and it is highly unbalanced, where 12 categories account for 50\% of the training samples.
We adopted this design to have a high coverage of object categories and avoid any bias towards any class.
With this set up, most of the test samples can be considered as out-of-domain from the perspective of the object category.
Figure~\ref{fig:product_type_histogram_by_train} details how the train partition is unbalanced, but the test partition is very balanced. 

% Category distribution of the train and test partitions

By default, AutoGluon uses the a partition 90\% of the provided data for training, and the remaining 10\% for internal validation.
The AutoGluon models were left to train as much time as needed, which was in the order of minutes for each binary classifier.
Models were trained with a p3.2xlarge cloud desktop in Amazon Web Services (AWS) equipped with a single Tesla V100-SXM2 GPU with 16 GB of memory.

Based on the results reported in Section~\ref{ssec:independent}, we consider three possible configurations: 1) a cascade of Autogluon binary classifiers (\textit{Cascade AG}), 2) a similar cascade where AutoGluon is replaced by Nova Pro for the \textit{product placement} defect (\textit{Cascade (AG \& NP)}), and 3) a single Autogluon binary classifier that does not distinguish between defect types (\textit{Single (AG only)}).
The performance of each configuration is reported in Table~\ref{tab:autoqa_sdxl}, together with the metrics of two set ups that facilitate the interpretation of results: 
(1) \textit{Random (0.5)} represents a lower bound baseline where the binary classifier would simply flip a coin, and (2) the \textit{Oracle} is the upper bound set by perfect predictions. %of would obtain perfect precision and recall metrics, and the ratio of labels would match teh ground truth.
%The \textit{No AutoQA} configuration refers to the case where no filtering at all is applied, so it corresponds to the situation where all images are considered as \textit{correct} with an inexisting AutoQA module. Notice how the precision for the correct clas in this case corresponds to the portion of correct images in the ground truth.

The AutoQA results presented in Table~\ref{tab:autoqa_sdxl} must be referred to the \textit{Random} baseline. 
A \textit{Random} AutoQA would have no effect on the quality of the generated images of the pipeline in Figure 1. Mathematically, this is equivalent to achieving a precision \precision~identical to the generator's one, $P_{\textrm{Gen}}$=\precision. 
In our use case, the quality of the generator is depicted in Table 4 as the yield of a perfect  \textit{Oracle} AutoQA is \ratio$=0.187$, . 
This value matches the precision \precision$=0.184$ of the \textit{Random} AutoQA.
The small difference is due to  $P_{\textrm{Gen}}$ being computed over the full 380 test images, but \precision over half (\ratio$=0.5$) the samples passed the \textit{Random} AutoQA.

Table~\ref{tab:autoqa_sdxl} shows that the three AutoQA configurations significantly improve over the \precision(\greentick) of the random baseline.
Similarly, their \precision(\redcross) may look high, but they are actually close to the random baseline because of the 81.3\% of \textit{Defect} cases in the data.
% where $r(\text{\greentick})$ is $0.19$ for the Oracle.

As proved in Equation~\ref{eq:autoqa_cost}, the total cost depends from both \ratio(\greentick) and \precision(\greentick), so we will focus on these two metrics.
When we do, we observe that using Nova Pro in the cascade does not offer gains over the AutoGluon only cascade.
Similarly to~\cite{wu2024q,he2024videoscore} we conclude that the considered off-the-shelf VLM does not provide reliable judgments for IQA, and that some additional adaptations would be needed, as proposed in~\cite{zhang2024quality,wu2024qinstruct}.
Between the two \textit{AutoGluon only} configurations, there is no clear winner when looking at Table~\ref{tab:autoqa_sdxl}. 
For this reason, we compare these two configurations in the next section. 
\begin{figure}[htb]
    \centering
    \includegraphics[width=0.45\textwidth]{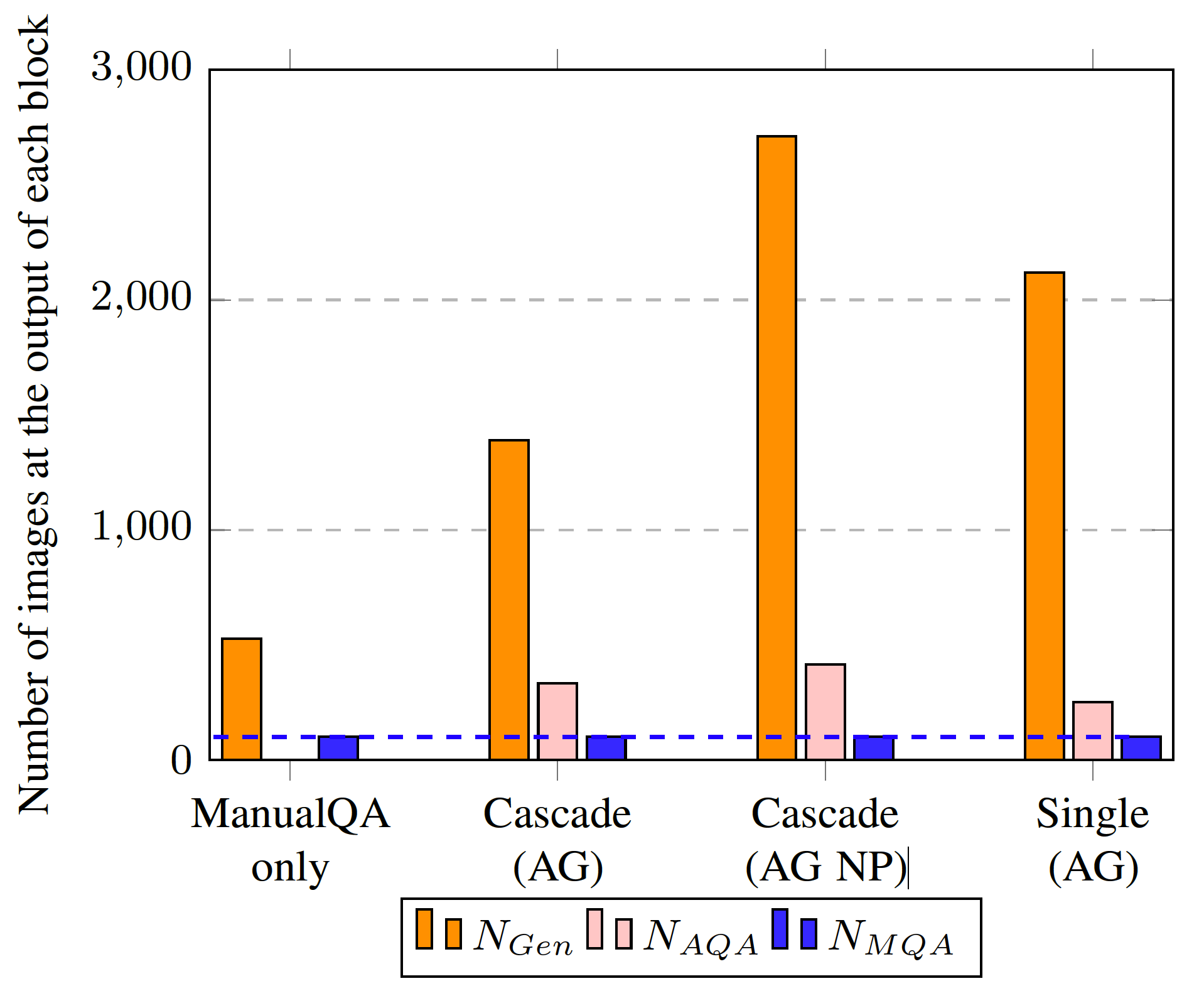}
    \caption{Volume of images needed after each block to obtain 100 high quality images.}
    \label{fig:volumes}
\end{figure}

\begin{figure}[htb]
    \centering
    \includegraphics[width=0.45\textwidth]{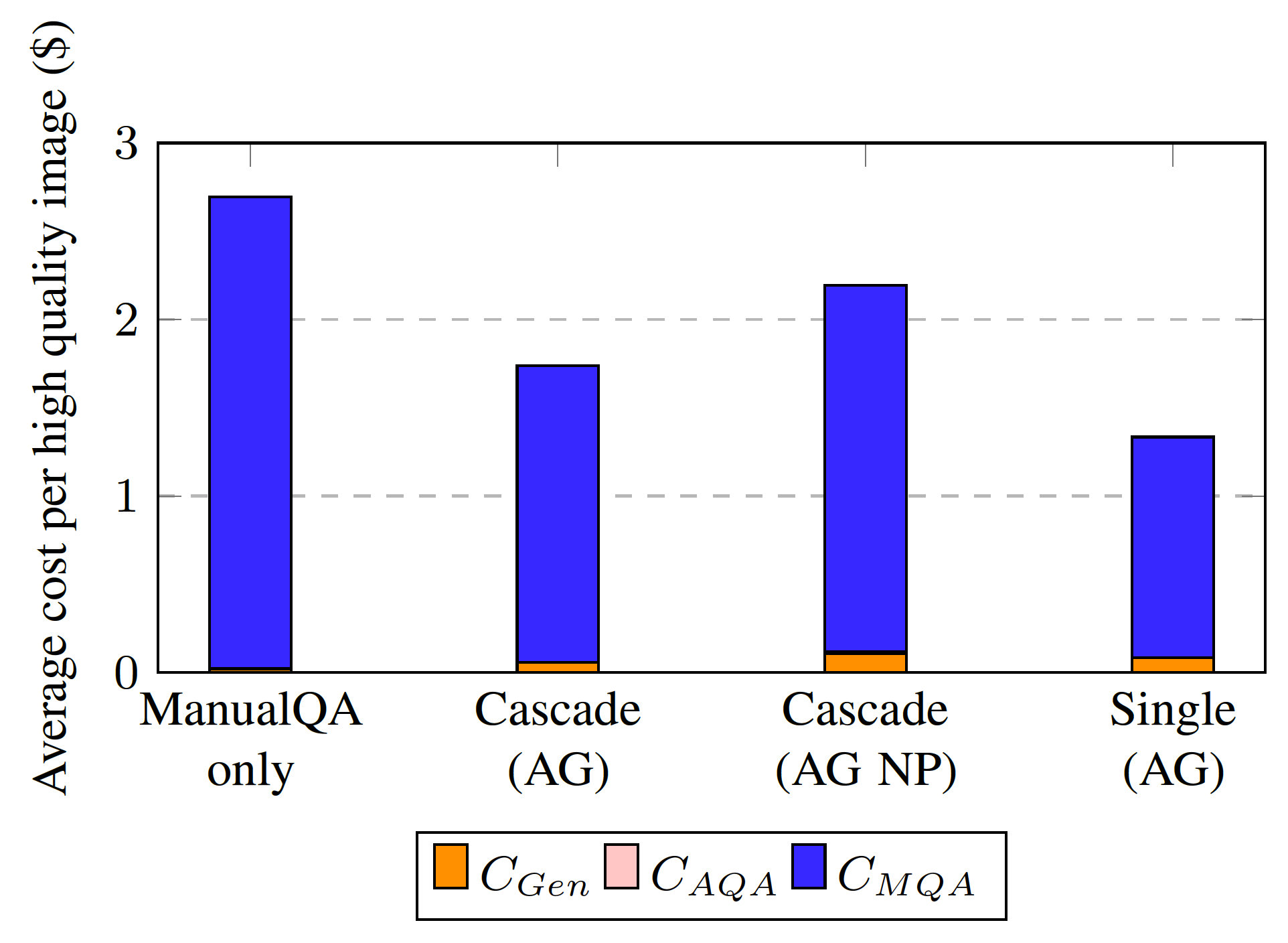}
    \caption{Cost composition over image generation ($C_{Gen}$), AutoQA ($C_{AQA}$), and ManualQA ($C_{MQA}$).}
    \label{fig:costs}
\end{figure}

\subsection{Cost savings}
\label{ssec:cost_savings}

The characterization of AutoQA configurations in terms of their \ratio(\text{\greentick}) and \precision(\text{\greentick}) allows estimating the cost savings, as presented in Equation~\ref{eq:cost_savings}. 
We use the measurements in Table~\ref{tab:autoqa_sdxl} to compare the three AutoQA configurations in our use case.

As a first step, we calculate the number of images needed at the output of each block in the pipeline, as developed in Section~\ref{sec:volume}. 
For simplicity, we consider a hypothetical case were the requirement is obtaining $N_{\MQA}=100$ high quality images. 
We first leverage Equation~\ref{eq:N_ManualQA_AutoQA} and \precision(\text{\greentick}) to obtain \Naqa, and afterwards Equation~\ref{eq:N_ManualQA_GenAI} and \ratio(\text{\greentick}) to estimate \Ngen~. 

Figure~\ref{fig:volumes} depicts the amount of \Ngen~images that must be generated, and the amount of \Naqa~images that must pass AutoQA, respectively.

The second step is weighting and aggregating the unitary cost of each stage with the volume of generated images, as presented in Equation~\ref{eq:cost_savings}. This requires establishing the unitary costs $c_i$, for $i \in \lbrace Gen, AQA, MQA \rbrace$:

% Generated images
\noindent \textbf{Image generation cost ($c_{Gen}$):} The unitary cost per generated lifestyle image is of \$0.00400 per image approximately, empirically estimated based on our experience of producing images on a cloud server. 

% AutoQA
\noindent \textbf{AutoQA cost ($c_{AQA}$):} 
The cost of running inference with AutoGluon is negligible compared to the rest of the costs.
However, the cost is not negligible if we consider one API call to Nova Pro on AWS Bedrock, which we approximate by \$0.00041 / image, assuming 3,000 input tokens and 300 output tokens.

% ManualQA
\noindent \textbf{ManualQA cost ($c_{MQA}$):} 
We follow the current suggestion in \textit{Amazon SageMaker Ground Truth pricing}~\footnote{\tiny\url{https://aws.amazon.com/sagemaker-ai/groundtruth/pricing/}} 
of $0.012$ for image classification tasks. 
We consider each of the 13 detailed defects defined as an image classification task, an we also multiply by 3, as we deal with three annotations per image. This makes \$0.468, which we round up to $c_{MQA}=\$0.5$. 
The reader is referred to~\cite{radsch2024quality} for discussion over the challenges of manual annotations.

The individual costs allow estimating the aggregated costs for the four considered set ups, 
as plotted in Figure~\ref{fig:costs}. 
The height of the bars show how the simple \textit{Single (AG)} is the cheapest configuration, offering a significant reduction of 51.61\% with respect to the \textit{ManualQA only} baseline.
Keep in mind that these high savings are estimated on a test set whose label distribution does not follow the training one. 
If both train and test partitions followed the same label distribution, we would expect to obtain a higher AutoQA precision, that would translate in even higher cost savings.
The composition of the plot bars show how the aggregated cost is clearly dominated by ManualQA, which represents 99.5\% of the total.

\begin{figure}[htb]
    \centering
    \includegraphics[width=0.45\textwidth]{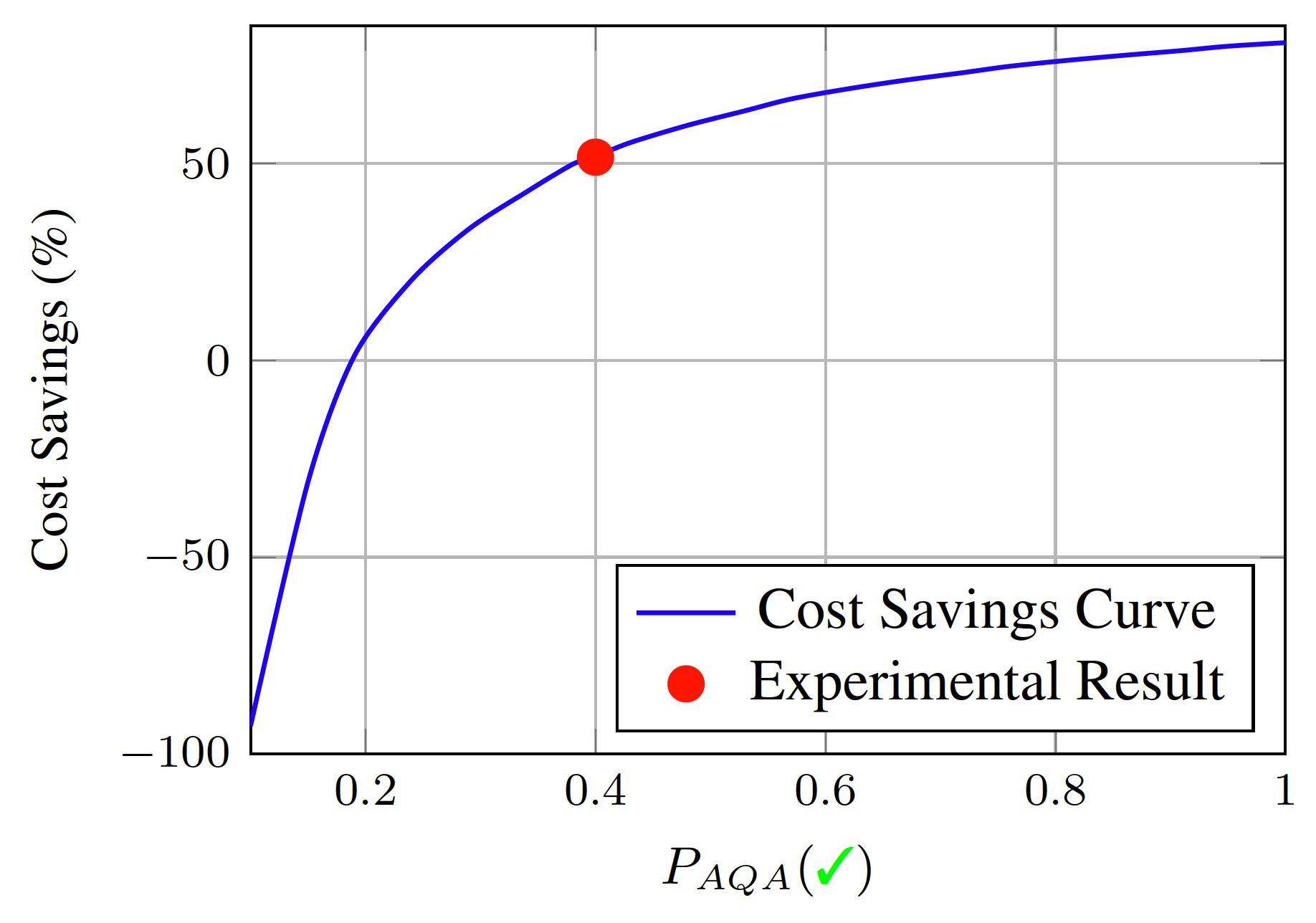}
    \caption{Cost savings as a function of AutoQA precision $P_{AQA}(\text{\greentick})$ with $P_{Gen}(\text{\greentick})$ set at 0.187, and $r_{AQA}(\text{\greentick})$ at 0.118. The red dot shows the experimental result from our Single (AG only) solution.}
    \label{fig:cost_sweep}
\end{figure}

Finally, we focus on the impact of \precision(\greentick), plotting how cost savings would evolve for other values beyond the $0.4$ of our use case.
The plot in Figure~\ref{fig:cost_sweep} shows how a very low \precision(\greentick) would make the AutoQA system harmful to a baseline without AutoQA, mostly because more images would need to be manually reviewed.
This effect is mostly represented by the last term of Equation~\ref{eq:cost_savings}, where the cost savings coming from ManualQA will become negative when \precision(\greentick) < $P_{Gen}$(\greentick).
Additionally, higher generation and AutoQA costs would also result from a low \precision(\greentick).

% python3 -c "
% import numpy as np

% def compute_cost_savings(N_MQA, r_AQA_clean, P_AQA_clean, P_Gen_clean, c_Gen, c_AQA, c_MQA):
%     if r_AQA_clean * P_AQA_clean == 0:
%         return 0  # Avoid division by zero
    
%     # First term
%     numerator1 = r_AQA_clean * P_AQA_clean - P_Gen_clean
%     denominator1 = r_AQA_clean * P_AQA_clean * P_Gen_clean
%     term1 = (numerator1 / denominator1) * c_Gen

%     # Second term
%     term2 = (1 / (r_AQA_clean * P_AQA_clean)) * c_AQA

%     # Third term
%     numerator3 = P_AQA_clean - P_Gen_clean
%     denominator3 = P_AQA_clean * P_Gen_clean
%     term3 = (numerator3 / denominator3) * c_MQA

%     # Final computation
%     result = N_MQA * (term1 - term2 + term3)
    
%     # Convert to percentage
%     baseline_cost = N_MQA * (c_Gen/P_Gen_clean + c_MQA/P_Gen_clean)
%     return (result/baseline_cost)*100 if baseline_cost > 0 else 0

% # Parameters
% N_MQA = 100
% r_AQA_clean = 0.118  # Fixed
% P_Gen_clean = 0.187
% c_Gen = 0.00209
% c_AQA = 0.0
% c_MQA = 0.5

% # Create 1D sweep
% p_values = np.linspace(0.1, 1.0, 20)

% print('% 1D Cost savings sweep data')
% for p in p_values:
%     savings = compute_cost_savings(N_MQA, r_AQA_clean, p, P_Gen_clean, c_Gen, c_AQA, c_MQA)
%     print(f'({p:.2f}, {savings:.1f})')
% "

% % 1D Cost savings sweep data
% (0.10, -92.8)
% (0.15, -30.8)
% (0.19, 1.0)
% (0.24, 20.4)
% (0.29, 33.4)
% (0.34, 42.8)
% (0.38, 49.8)
% (0.43, 55.3)
% (0.48, 59.7)
% (0.53, 63.4)
% (0.57, 66.4)
% (0.62, 69.0)
% (0.67, 71.2)
% (0.72, 73.1)
% (0.76, 74.7)
% (0.81, 76.2)
% (0.86, 77.5)
% (0.91, 78.7)
% (0.95, 79.8)
% (1.00, 80.7)

\begin{figure}[!htb]
    \centering
    \includegraphics[width=0.4\textwidth]{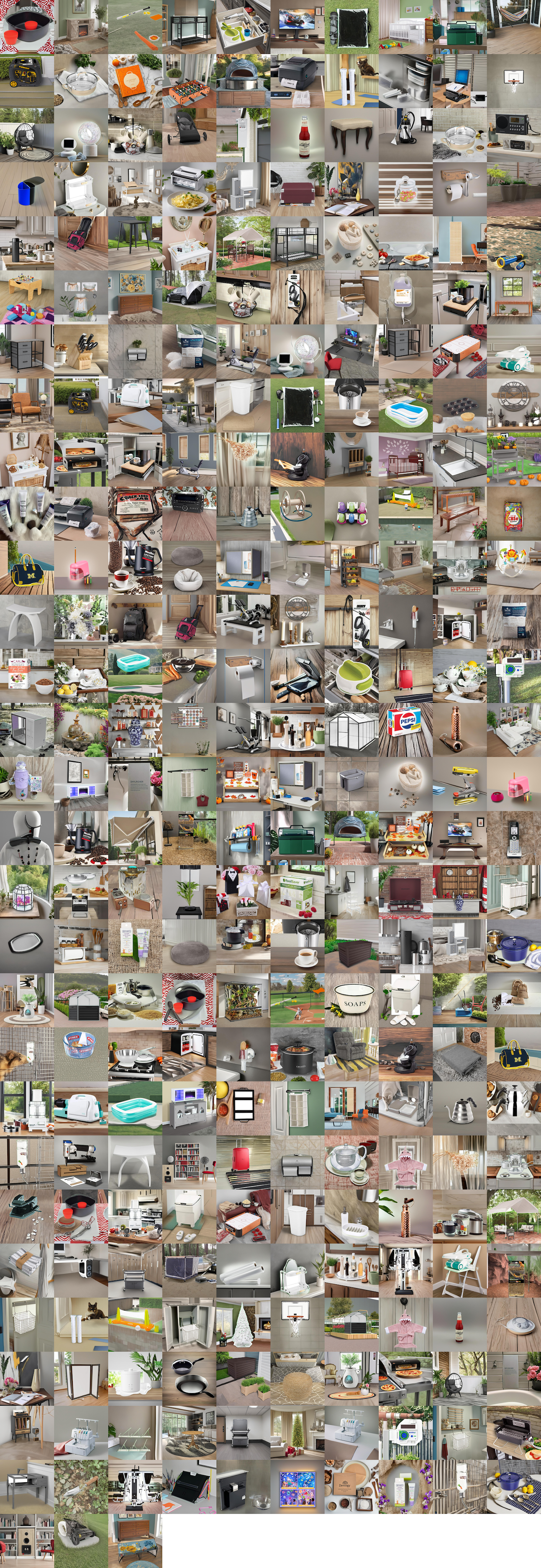}
    \caption{Full set of True Defect (\text{\redcross}) predictions.}
    \label{fig:tp_qualitative}
\end{figure}

\begin{figure}[!htb]
    \centering
    \includegraphics[width=0.4\textwidth]{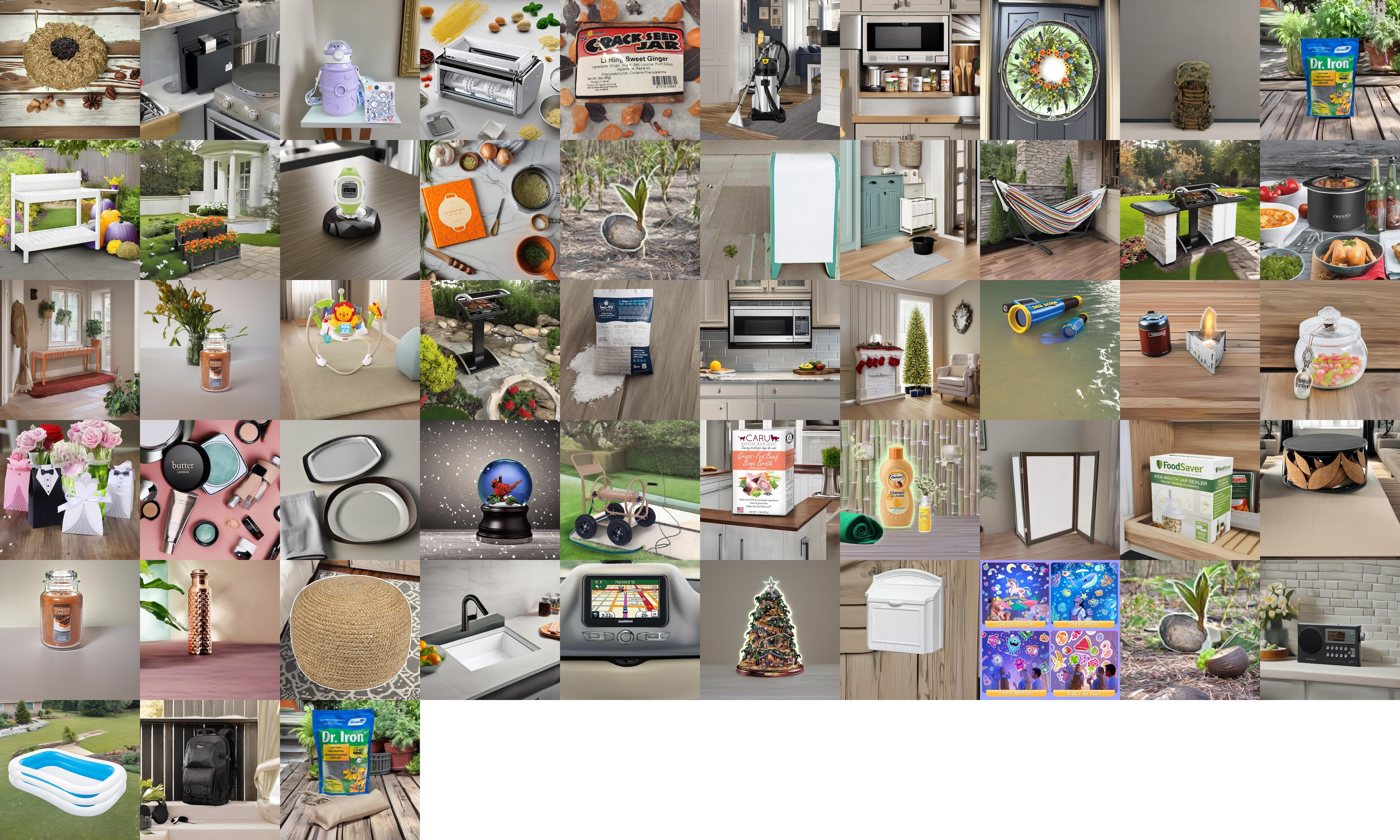}
    \caption{Full set of False Defect (\text{\redcross}) predictions.}
    \label{fig:fp_qualitative}
\end{figure}

\begin{figure}[!htb]
    \centering
    \includegraphics[width=0.4\textwidth]{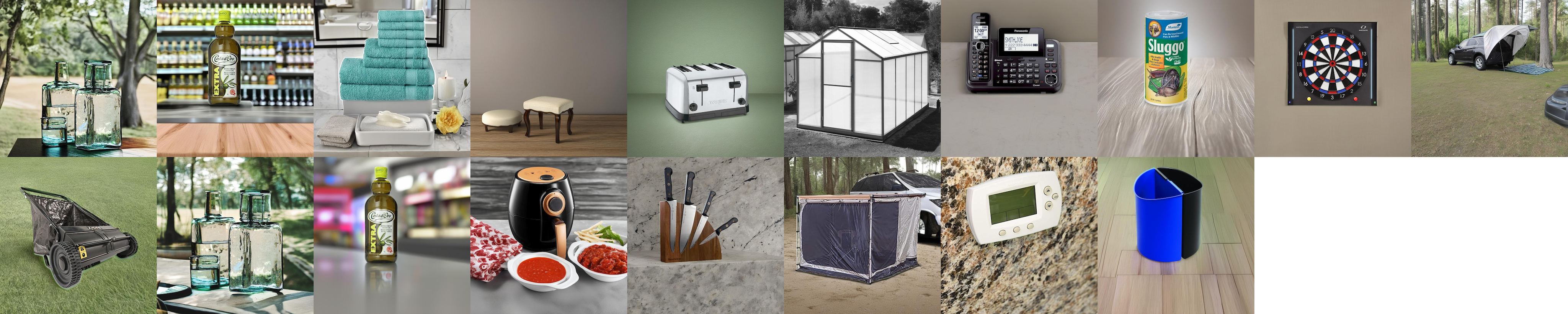}
    \caption{Full set of True Clean (\text{\greentick}) predictions.}
    \label{fig:tn_qualitative}
\end{figure}

\begin{figure}[!htb]
    \centering
    \includegraphics[width=0.4\textwidth]{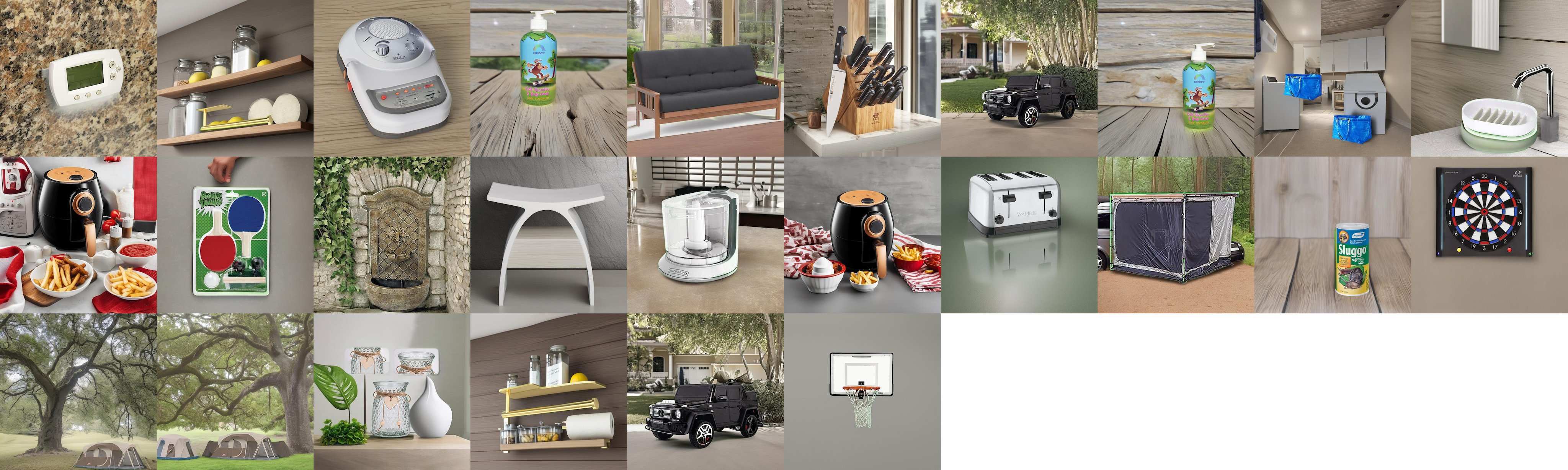}
    \caption{Full set of False Clean (\text{\greentick}) predictions.}
    \label{fig:fn_qualitative}
\end{figure}

\subsection{Qualitative Results}
\label{ssec:qualitative}

includes The full predictions in the test set are provided in Figures~\ref{fig:tp_qualitative}-\ref{fig:fn_qualitative}. A visual inspection of the data does not show evidences of biases towards certain objects or backgrounds.

In the early stages of our research, we did observe that the VLM-based AutoQA was biased to only allow images with very simple backgrounds, which were less valuable for our target e-commerce application. For this reason, we extended the prompt to include the detailed defect \textit{Rich background} shown in Table \ref{tab:defects_prompts}. When we experimented with AutoGluon we no longer observed this issue. 

\graphicspath{{figures/qualitative/}}

% \begin{figure*}[!htb]
%     \centering
    
%     % True Clean Images (1x4 grid)
%     \subfloat[True Clean (\text{\greentick}) predictions.]{
%     \includegraphics[width=\textwidth]{./tp_grid.png} 
%     }\quad

%     % False Clean Images (1x4 grid)
%     \subfloat[False Clean (\text{\greentick}) predictions.]{
%     \includegraphics[width=\textwidth]{./fp_grid.jpg} 
%     }    

%     % True Defect Images (2x5 grid)
%     \subfloat[True Defect (\text{\redcross}) predictions.]{
%     \includegraphics[width=\textwidth]{./tn_grid.png}     
%     }\quad    
        
%     % False Defect Images (2x5 grid)
%     \subfloat[False Defect (\text{\redcross}) predictions.]{
%     \includegraphics[width=\textwidth]{./fn_grid.png}     
%     }
    
%     \caption{All AutoQA predictions provided by the Single (AG only) configuration on images generated with an SDXL GenAI model.}
%     \label{fig:qualitative}
% \end{figure*}

\section{Conclusions}

We have shown how introducing AutoQA to an image generation pipeline can bring significant cost savings.
The formula derived estimates these savings based on the AutoQA yield and precision.
While we have applied this formula to the use case of image generation, it is valid for any GenAI task.

Our study case for background inpainting has shown significant cost savings of 51.61\% even with a modest AutoQA precision of 0.4.
This is because the cost of ManualQA clearly dominates over the costs of the automatic blocks of the pipeline.
AutoQA comes almost for free, and it increases the quality of the images sent for AutoQA. In our best configuration, 
AutoQA only approved 11.8\% of the generated images but, at the same time, 
the ManualQA yield increased from 18.7\% to 40.0\%.
As a consequence, annotators need to review less images to reach a certain goal, and the total cost decreases. 

The proposed technical solution is simple and based in AutoML, which allows grounds for improvement based on the existing literature on IQA.
In our set up, the zero-shot VLMs mostly could not detect the defects in the images, a limitation aligned with existing works that question the effectiveness of this set up for VQA problems~\cite{ross2024what} and image quality assessment~\cite{huang2024aesbench,he2024videoscore,zhang2024q}.
A fine-tuning of these models should be explored to improve their performance.

The significant cost saving achieved in our work motivates further scientific research. 
This research can be oriented in two directions: improving the image generator or the AutoQA engine. 
In any case, the ultimate goal is completely removing the need of a manual review in the pipeline. 

{
    \small
    \bibliographystyle{ieeenat_fullname}
    \bibliography{main}
}

\end{document}